# AI-assisted pre-review of open-source software submissions: an experience report from BOSC 2026

Authors: Tazro Ohta[1,*], Nomi L. Harris[2], Seth Carbon[2]

1. Department of Artificial Intelligence Medicine, Graduate School of Medicine, Chiba University, Chiba, JAPAN
2. Lawrence Berkeley National Laboratory, Berkeley, CA, USA

* Corresponding author, <tazro.ohta@chiba-u.jp>

## Abstract

Most conferences rely on peer-review of submissions, but as generative AI makes it easier than ever to prepare submission materials, some conferences are seeing an overwhelming surge of submissions. We wanted to see if generative AI could help our conference's volunteer reviewers by pre-reviewing abstracts for certain criteria. The Bioinformatics Open Source Conference (BOSC) was well-positioned to experiment with this, as we already had a detailed rubric used by reviewers to evaluate submitted abstracts on multiple criteria, including openness (public availability of the code or other content associated with the project), valid open source license, and "runnability" (how easy it is to download, build, and run the project - an important measure of reusability). For BOSC 2026, we built bosc-pre-review, an agentic skill that assessed six review criteria, and Runabilly, which builds and tests each project in a disposable Docker container for safety. The AI only gathered evidence to present to the reviewers; humans made every decision regarding the acceptance of the abstracts. After the review period, we surveyed the reviewers to determine how useful they found the pre-review. Most of those who responded said they found it useful, but they preferred to check the AI's conclusions against their own, rather than accepting the AI results unquestioningly.

### Keywords

AI-assisted peer review; conference review; open-source software; reproducibility; bioinformatics; large language models

# 1. Introduction

The rapid advance of generative AI has made researchers dramatically more efficient, accelerating both how they take in information and how they produce results. Yet this anticipated acceleration of science brings a separate and growing challenge: ensuring that what is produced is actually valid. Increasingly, work created with the help of AI is submitted as a research output — to journals and to conferences — without having been adequately verified even by the people who generated it [1]. The review of such submissions has always rested on the volunteered effort of the research community, and that community now faces pressure to adapt to the shifts in both the quality and the quantity of submissions that AI has set in motion.

The Bioinformatics Open Source Conference (BOSC, open-bio.org/events/bosc), held yearly since 2000, is a meeting for researchers and developers in bioinformatics focusing on transparency in science through the promotion of open source and open data. Publishing and sharing data and methods does more than improve the reusability of data and software: it underpins reproducibility, a foundational principle of science. At the same time, transparency is itself a challenge for researchers. As the emergence of the FAIR principles showed, simply releasing data and code is not enough to make them genuinely usable [2]. BOSC has long worked to show the research community what good sharing looks like, examining the licensing of data and code, the quality of documentation, and the reusability and reproducibility of the software behind each submission. Like the peer review of journal articles, however, BOSC depends on review carried out by volunteer researchers and developers. This is important work that deserves recognition, but it is also a burden that seldom directly advances the reviewer's own career.

BOSC has had an openly shared review process and rubric for years [3], and we are always on the lookout for ways to improve its efficiency and fairness. For BOSC 2026, we introduced a generative-AI-assisted pre-review intended to lighten the load on reviewers. The guiding principle was that human reviewers would always make the final decisions; AI was used only to gather, in advance, the facts relevant to each submission and make these available to the reviewers, so that reviewers could focus on using their expert judgment rather than fact-finding. This approach differs from other reported approaches, which either replace human reviewers with AI or use AI reviews alongside human ones to reach a consensus [4]. This paper reports in detail on that experimental AI-assisted pre-review process.

# 2. BOSC review

## 2.1 Human review

BOSC is an entirely volunteer-run meeting that has been held most years as a track of the big Intelligent Systems for Molecular Biology (ISMB) conference. The BOSC Organizing Committee recruits volunteer reviewers from the bioinformatics open source community. Most years, there are about 25 reviewers, and between 60-100 abstracts submitted. Abstract submission and review is done in the EasyChair platform. To help reviewers gauge the openness and reusability of submitted work, the BOSC submission form requests, in addition to a 1-2 page abstract, answers to some questions such as “What is the URL of your open content repository?” and “What is the license?”.

In the review process, each abstract is assigned to three reviewers, who follow our review process and rubric [3]. Each field in the review form asks the reviewer to rate a specific aspect of the submission, and, optionally, supply comments (Fig. 1). The overall “score” for an abstract, which is used to help decide which abstracts to offer talks or posters to, is calculated from a weighted combination of the reviewers’ assessments of the individual criteria. Some criteria (e.g., content is openly available at the provided URL) are weighted more heavily than others (e.g., “Novelty”).

**Available.** Did the author specify a URL for their code / data / materials (e.g., documentation), and is it in fact accessible at the listed URL?

- yes
- not sure / not applicable (explain)
- no

**Figure 1.** The form used by (human) reviewers to rate BOSC abstracts has a series of review criteria with multiple choice answers and free-text comment boxes. Above is one example. The full list of criteria can be found at https://github.com/OBF/bosc_materials/blob/master/BOSC_review_process.md.

BOSC reviewers typically do an impressively thorough job of assessing the abstracts. Some questions can be answered quickly (e.g., is the content accessible at the supplied URL?), but others take longer. For example, one optional question asks how easy the material is to access, install, and use (“runnability”). Many reviewers do choose to answer this question, which (for code) involves downloading the code, possibly building it, and trying to run it. This can take many minutes or even hours. Authors have told us that they appreciate the detailed comments that our reviewers provide, but not all reviewers have the time or background information to do these assessments.

## 2.2 The BOSC pre-review model

Our experiments with AI-based pre-review used the model where AI-generated assessments could be viewed by the human reviewers, and compared with the reviewers' own expert opinions. There was no intention of replacing humans with AI in the review process; our goal was to provide the reviewers with information to make the review process easier and more reliable. Pre-review results were shared with reviewers in the form of a spreadsheet. AI-generated scores were not used in calculating the total score for each abstract.

For our pre-review experiment, we focused on criteria that are more objective (such as "Available") and/or more time-consuming for human reviewers (such as "Runnable"). The criteria we chose, and the scoring rubric for each, are shown in Table 1 and described in section 3.1. In addition to runnability, another question that was not easy for reviewers to answer was, "Has this project been presented previously at a recent BOSC, and, if so, does the new abstract present significant updates?" (this criterion was called "Fresh" for short). Additionally, questions that are basically objective (like whether the supplied license is a valid open-source license and is found in the repository) can potentially be answered by automated approaches. Questions like these seemed like excellent targets for semi-automation, which motivated us to experiment with using generative AI to do a "pre-review". Meanwhile, questions that demand more expert knowledge, experience, and subjective opinion – such as whether an abstract is relevant to BOSC, or whether it lays out a plan for sustaining a user community – may not be as suitable for automation.

# 3. System design

## 3.1 Pre-reviewer architecture and process

To implement our pre-reviewer ("bosc-pre-review") [5], we used Claude Code, Anthropic's command-line coding agent, but the overall process is agnostic to the specific AI tool used. Bosc-pre-review reads each submitted abstract and associated metadata and scores it against BOSC's rubric. It is implemented as a skill; in Anthropic's terminology, a *skill* is a set of instructions and scoring criteria that the agent loads and follows when it is invoked. In addition to instructing Claude, skills are human-readable and human-modifiable.

The input to bosc-pre-review is a tab-delimited table extracted from the conference's EasyChair export, with one row per abstract. Each row includes the abstract's identifier, its title, the license the author declared on the submission form, and the source URL the author provided for their open content. The full text of the abstracts was also made available to the agent. For each abstract, the agent visits the source URL and inspects whatever repository or page it finds there: the file listing, the README, any license file, and, for code hosted on GitHub, the commit history and contributor counts from the GitHub API. It then assigns a score for each criterion and writes a one-line comment explaining it. The output of the pre-review is a tab-delimited table of scores and comments, one row per abstract, ready to paste straight into the spreadsheet shared with reviewers.

The pre-reviewer assessed six criteria, each on a scale that maps to the answer choices on the review form that reviewers were provided with (Table 1). “Available” asks whether the content (whether source code or other materials) is actually retrievable from the specified URL. “Open” checks whether the open repository includes a recognized open-source or open-content license that matches the author's declaration. “Fresh” asks whether the same project, or a closely related one, has been presented at BOSC in the past four years (the past four years’ abstracts were supplied to the AI agent). “Active” asks whether the project is actively maintained. “Runnable” (described in more detail in the next subsection) asks whether the code can be built and run easily.

We added an experimental field, “Human-generated”, that is not on the review form. We shared the results with the reviewers only to help them put the submission in context. The possible results for this field are "looks human", "some AI assist", or "likely vibe-coded", the last meaning a project that appears to have been produced largely by an AI agent with little human authorship. To assess this, Claude looks in the repository associated with the submitted project for signs such as AI tool-configuration files (for example .claude/, AGENTS.md, or CLAUDE.md), commit histories made up of formulaic bursts, commits that strip out earlier references to AI tools (which we actually found in one submission), README content in a characteristically generated style, and implausibly large output from a single contributor over a very short period. In line with ISCB's policy on the use of generative AI, BOSC encourages its appropriate use and does not treat AI-assisted development as grounds for rejection [6]. This field therefore only informs reviewers and is not used to penalize AI use.

**Table 1.** The bosc-pre-review rubric. Each abstract is scored on six criteria, each tier mapping to a multiple-choice answer on the EasyChair review form.

| Criterion | Top | Medium | Low | Failing | Not applicable |
|---|---|---|---|---|---|
| Available | Accessible URL that points to the correct source repository | — | Some content present, but the complete/correct source is not clearly there, or the project is not identifiable within a large repo/org | Unreachable, empty, or website-only with no findable source repository | Non-software abstract |
| Open | Valid open-source license, file present, matches the declaration | Valid license, but the license file is missing or does not match | Not a recognized open-source license | No license, or source unavailable | Non-software abstract |
| Fresh | New to BOSC (no prior appearance, 2022 to 2025) | Prior appearance, clearly different aspect | Prior appearance, incremental update | Essentially identical resubmission | — |
| Active | Multiple contributors, good docs, active issue tracker | ≥2 contributors, active in the past year, >2 weeks old | Single contributor, short burst, or stale >1 year | No accessible source | Non-software abstract |
| Runnable | SUCCESS — builds and the project's tests pass | WARNING — builds, but full test validation is blocked by | — | FAILURE — no working build: build | UNDEFINED — repository holds no |
|  |  | an environmental hurdle |  | fails after retries, tests fail, 1-hour timeout, declared URL not found, or code not packaged as a buildable project | software to build (e.g. training materials, docs, dataset) |
| Human-generated | no signal — nothing in the repo suggests AI generation | some AI assist — normal AI usage visible (e.g. commits mention AI tools) | likely vibe-coded — multiple strong red flags | source unavailable — can't assess | can't assess — source unavailable |

## 3.2 Runabilly

One of the criteria that BOSC asks reviewers to check–which few other conferences would look at–is how easy the project described in an abstract is to access, install, and use (its "runnability"). This is directly relevant to the reproducibility of the work: if other researchers can't build or run your code, then they can't reproduce your results. However, although this is an important requirement for open-source projects, it is also time-consuming and challenging for reviewers to evaluate, so assessing runnability was one of our top goals for the BOSC pre-reviewer.

Runabilly [7] is an experimental automated system to determine whether an open-source application, package, or library is practically buildable and usable from a given source repository. The Runabilly stack spins up a disposable Docker container (for safety as well as convenience), clones an open source project into it, and uses Claude (or similarly functional agent) to automatically explore, install dependencies, build, and report the results. Runabilly uses a minimal Ubuntu 24.04 base image with only basic tools (git, git-lfs, curl, build-essential, etc.). No language-specific toolchains are pre-installed--they get added as needed for each project.

The product of a Runabilly run is a two-part structured evaluation. The first section is a "build result":

- SUCCESS — build completes AND the project's tests run and pass (or no test infrastructure is present). Tests are required when present: a passing build with failing tests is not a SUCCESS.
- WARNING — build completes but full test validation is blocked by an unavoidable environmental hurdle (e.g. Docker-in-Docker, large external databases, paid API keys, GPU-only). The code looks healthy; the environment can't validate it. The specific hurdle is named in the report.
- FAILURE — build fails after retries, OR tests run but fail, OR the 1-hour timeout is exceeded, OR the URL is not found
- UNDEFINED — URL isn't a buildable repo (e.g. training materials, landing page, documentation site, dataset collection)

The second output from Runabilly is a “difficulty rating”, trying to capture how hard it would be for a human to build the project. The main criteria were time to build, number of dependencies, how “exotic” the build was, and how much the documented build diverged from the build as it was actually run.

**Table 2.** Runabilly “difficulty” rubric, quantifying how hard it would be for a human to build the project.

| **Factor / Difficulty** | **Low** | **Medium** | **High** |
|---|---|---|---|
| Time | < 1m | 1m - 5m | > 5m |
| Dependencies | < 10 packages | 10 - 50 packages | > 50 or multiple toolchains |
| Exoticness | Standard build system, no workarounds | Less common build system or minor workarounds | Custom scripts, multi-stage setup, Docker-in-Docker, etc. |
| Divergence | Documented build path worked on first try | Minor adjustments needed | Documented path failed; alternate route required, or no docs |

A quick roll-up was also provided: EASY (all factors LOW), MODERATE (any MEDIUM, no HIGH), HARD (any HIGH), IMPRACTICAL (can't complete in a disposable container).

The approach used by Runabilly, building software in an isolated environment such as a Docker container, not only reduces failures caused by dependency issues but also helps maintain the security of the testing environment. Downloading and installing software from external sources can pose security risks, regardless of the developer’s intentions. Developers should therefore take reasonable steps to assess and minimize such risks before release. In this regard, Runabilly may be useful not only for BOSC reviews but also in broader contexts where software needs to be tested safely and reproducibly.

## 3.3 Deployment at BOSC 2026

For BOSC 2026 we ran our pre-review workflow and shared the results with reviewers in the form of a spreadsheet. The workflow started with exporting the submissions from EasyChair with the full abstract text included, removing the irrelevant or sensitive columns, and passing them to bosc-pre-review. Every submission whose source was a code-bearing GitHub repository then went on to Runabilly, which tried to build and test it in a disposable container. Submissions whose "source" was an organization page, a list of several URLs, a website, or a non-software artifact were marked not applicable for the build step. Abstracts submitted to BOSC cover topics that include not only software with publicly available source code, but also open data projects that make datasets and databases publicly available, as well as discussions related to bioinformatics education or policy. The ‘Runnability’ status for these submissions is treated as ‘Not applicable’ and is clearly distinguished from ‘failed’.

The output of the pre-review workflow was treated as advisory information throughout the review process. Reviewers were free to consult the pre-review results, which were shared separately from the review system, and made every accept, reject, talk or poster decision themselves. The AI-generated scores never entered the calculation of an abstract's total score. Because the runnability check was the most time-consuming thing a reviewer could do by hand, reviewers were asked not to repeat it with their own AI tools, though they were welcome to form their own judgment on the question. By taking on the tedious, mechanical parts of the review, the pipeline let reviewers spend their time on the parts that benefit from human expertise.

# 4. Results

## 4.1. Calibration and tuning

Before the production run, we compared the automated criteria with human judgments and adjusted our skill where the two disagreed. The benchmark was a set of abstracts from past BOSCs that had already been reviewed and whose accept or reject outcomes were already known. Each disagreement between the human and AI assessments pointed to a place where the rubric was underspecified, and each led to a concrete change that we built into bosc-pre-review. For example, when assessing “Active”, a repository built intensively by several people over a short period and then left untouched fell into the gap between tiers. For another abstract, activity counted at the level of a single repository diverged from the activity visible across the parent organization. For “fresh”, a project resubmitted under a different name slipped past the title matching check. And for “available”, a URL that resolved to a large monorepo was scored as available even though the specific project could not be identified inside it. In response to each of these, we adjusted the pre-reviewer. “Fresh” now matches on author and full text, “Available” confirms that the project and the repository correspond. For the “Active” field, hackathon-style repositories get a specified treatment, and the parent organization's activity is taken into account.

## 4.2. Results from the 2026 run

We ran our pre-reviewer on 60 submissions to BOSC 2026. Of the 60, 39 were found to have an available repository and 37 had a license that was present and matched the author’s declaration. Only a few clearly failed these important criteria: 3 had unreachable or nonexistent repositories, and 3 had no usable license. On the “Fresh” criterion, the majority (50 of 60) were new to BOSC, with 10 prior matches. “Active” was the most spread-out criterion and skewed low: only 20 of 60 reached the top tier, and 16 were single-contributor or short-lived repositories. The advisory Human-generated note came out as 45 "looks human", 11 "some AI assist", and 4 "likely vibe-coded".

The low average scores for “Active” are not a scoring fault but a reflection of the submission pool. Many 2026 BOSC submissions represented small, new projects built mostly by one person. This is not necessarily a flaw, and this score was not weighted heavily in the total score, but was occasionally used as a tie-breaker. The license problems took a few forms: a license declared on the form but absent from the repository, a license that is not OSI-approved, or a mismatch between a license's official name and the way the repository actually labels it. When we looked at the “Human-generated” results, we saw, as we had expected, heavily AI-dependent development in a few submissions, including one project that had deliberately removed the AI signatures.

The “Fresh” check was useful mainly for spotting submissions that report updates to software or platforms presented at BOSC before. Until now, human reviewers had done only a quick check, typically web-searching the title to see whether a project had appeared at BOSC previously. By examining the full text of the abstract, the “Fresh” check correctly identified presentations on projects that had already been presented recently at BOSC, even if the abstract title was substantially different. Because BOSC values long-maintained open-source software and systems, it does not assess project update submissions negatively; it is just helpful for reviewers to know whether a project is new to BOSC or has been presented previously.

Runabilly was a useful time-saver for checking whether a project builds and runs. For projects where it could find a code repository, Runabilly attempted to build and run it. Of the 60 submissions, it attempted a build for all but 16. Of those 16, 4 submissions had URLs that did not point to source code repositories, and 2 were duplicate submissions, so were not tested. The remaining 10 were cases where the submitted URL did not resolve to a single buildable repository, but most of these were not deficiencies; they just reflected the structure of the project's codebase. For 5 submissions, the URL pointed to a relevant organization or set of repositories that held the project's real code, just not in a single buildable form. Another 2 were policy or ecosystem contributions that point to a relevant community but were never meant to be a single buildable artifact. Two more were borderline: the URL was available and plausibly relevant, but the specific artifact described in the abstract, such as a benchmark, was not present in the repository itself. Only one submission pointed to a reachable website with no relevant source code at all. Of the 44 that Runabilly did attempt, 17 built and passed their tests, and 18 returned a warning. A warning means the code looked healthy, but the disposable container could not supply something it needed, such as a GPU, a commercial license, large reference data, paid API keys, or a prerequisite database. Another 5 were recorded as failures, meaning Runabilly ran but could not produce a working build: 1 was a genuine source-code problem (broken tests); 2 were URLs that returned a 404; and 2 were code that was not packaged as a buildable project (workflow definitions or a Colab notebook). The last 4 were classified as "undefined": Runabilly inspected the repository but found no software to build, only training materials or data. The high proportion of warnings is the main lesson. Much of the code submitted to BOSC is not broken, but a lot of it cannot be fully exercised outside the specialized environment it depends on. Such code is not fully "FAIR", as it fails the Reusability principle, so it should not receive a high score for runnability, but will not usually be rejected from BOSC for that reason alone.

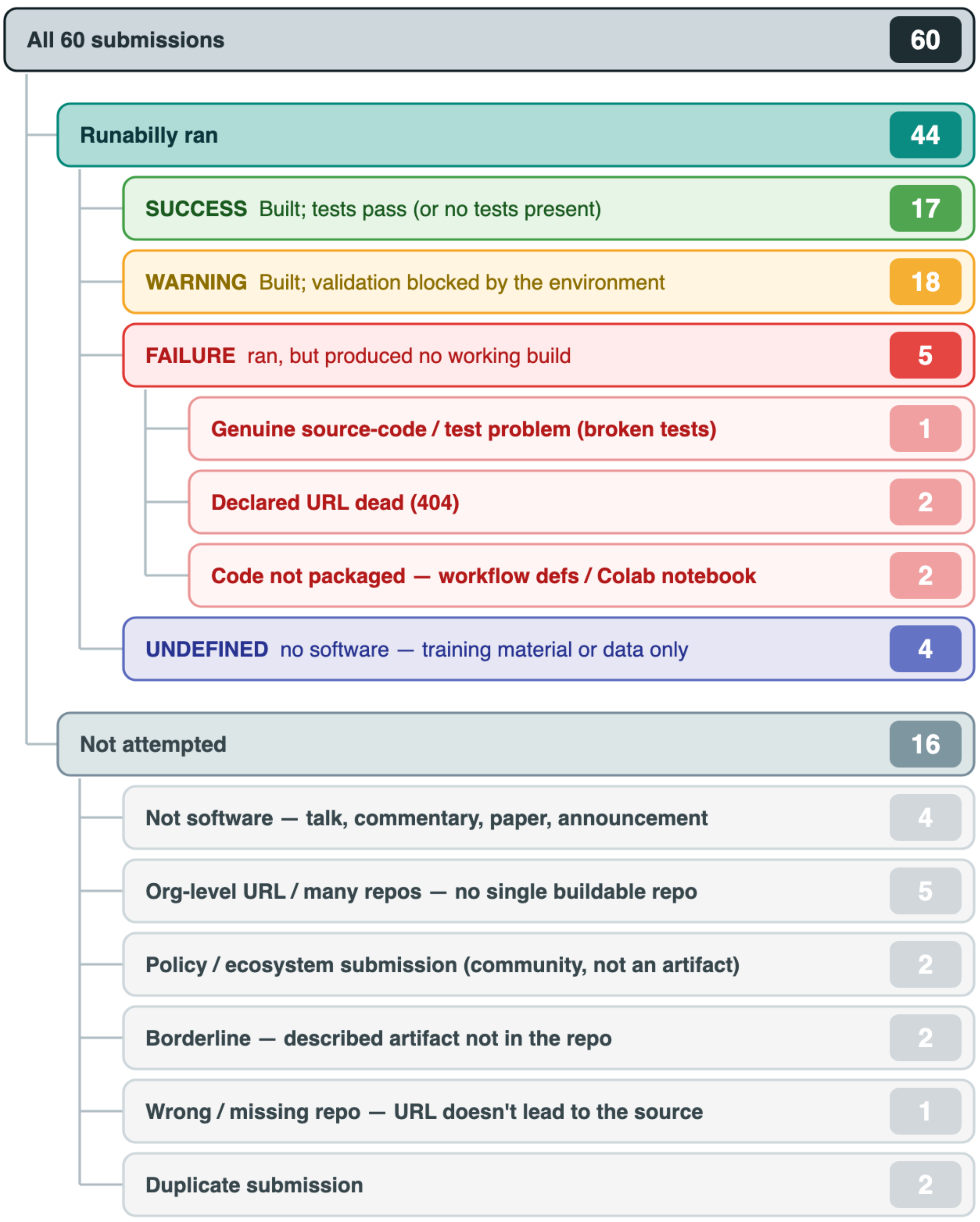

**Figure 2.** Runabilly [7] outcomes for all 60 BOSC 2026 submission records. Each record is routed by whether a single buildable repository could be found (44 attempted, 16 not), then by build result. WARNING = code looked healthy but the disposable container lacked something it needed (GPU, commercial license, large reference data, paid API keys, or a prerequisite database). FAILURE =

Runabilly ran but produced no working build (broken build/tests, a dead URL, or code not packaged as a buildable project). UNDEFINED = the repository was inspected but held no software to build (training material or data only).

## 4.3 Reviewer feedback

After the review process was finished, we sent the 28 reviewers a survey (Appendix A) to determine how useful they found the pre-reviewer. Seventeen reviewers responded. Of these, 15 had looked at the pre-review results, 3 of them closely and the rest somewhat, while 2 had not looked at all. Of the 15 who used the results, 5 found them very helpful and 10 somewhat helpful, and none found them unhelpful. Most estimated that the pre-review saved them a few minutes per review (Fig. 3).

How much time do you estimate the pre-review saved you, compared with the time you would have spent without it?
15 responses

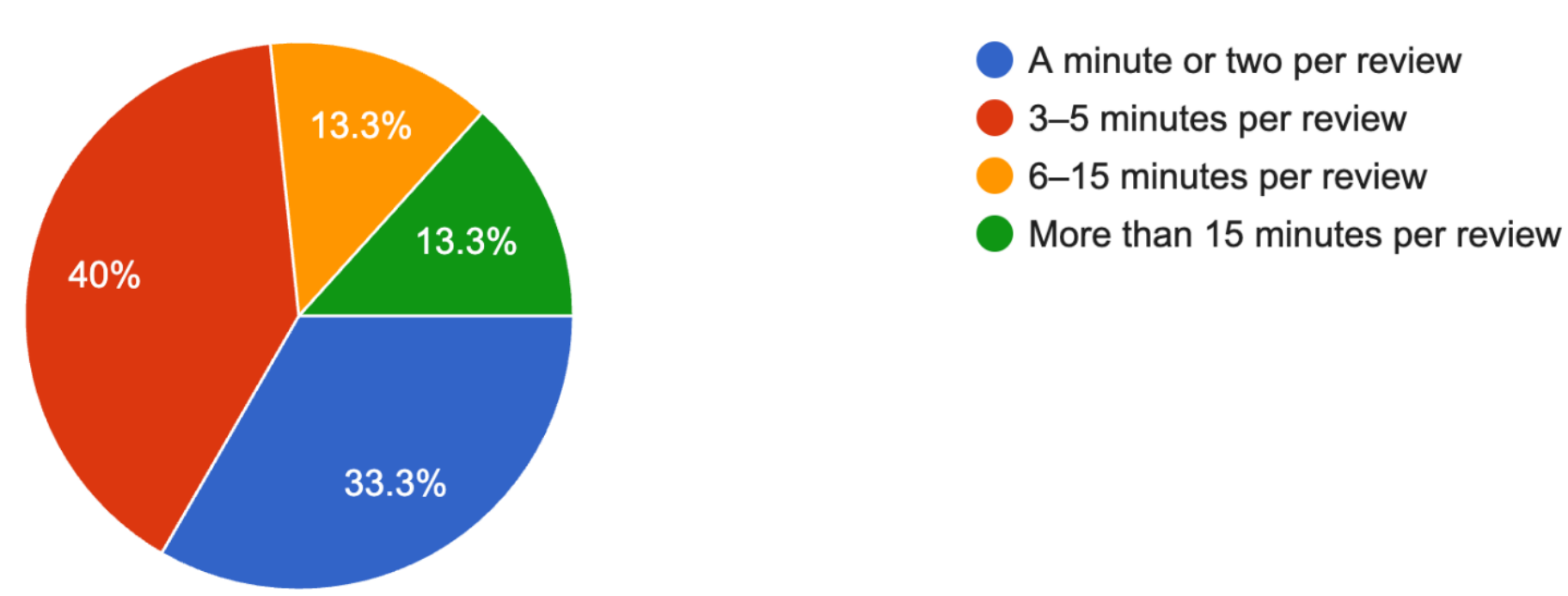

**Figure 3.** 10 out of 15 reviewers who responded to the survey said that access to the pre-review results saved them at least 3 minutes per review, with two saying it saved them more than 15 minutes per review.

Reviewers found the pre-review results for the more objective criteria the most useful. “Availability” and “license” were each cited as helpful by 13 of the 15, freshness by 12, and activity by 10. “Runnability” (which was an optional review field) was the criterion reviewers were least likely to check themselves: 10 did not assess runnability at all, 9 tested at least some builds by hand, and 4 relied on the Runabilly results.

Trust in the pre-review results was moderate, with 6 of the 15 reviewers saying they mostly trusted the results and 9 somewhat; none reported complete trust or none at all (Fig. 4). Their verification habits matched this caution: 9 always double-checked the AI-generated assessments and 6 did so sometimes. Only 2 reviewers reported noticing any wrong or misleading output, both minor. One was a borderline Runabilly rating that the reviewer rechecked by hand and judged close enough. The other was a license read from a shallow fork whose license differed from the original project, a case the reviewer noted is hard for humans to get right as well.

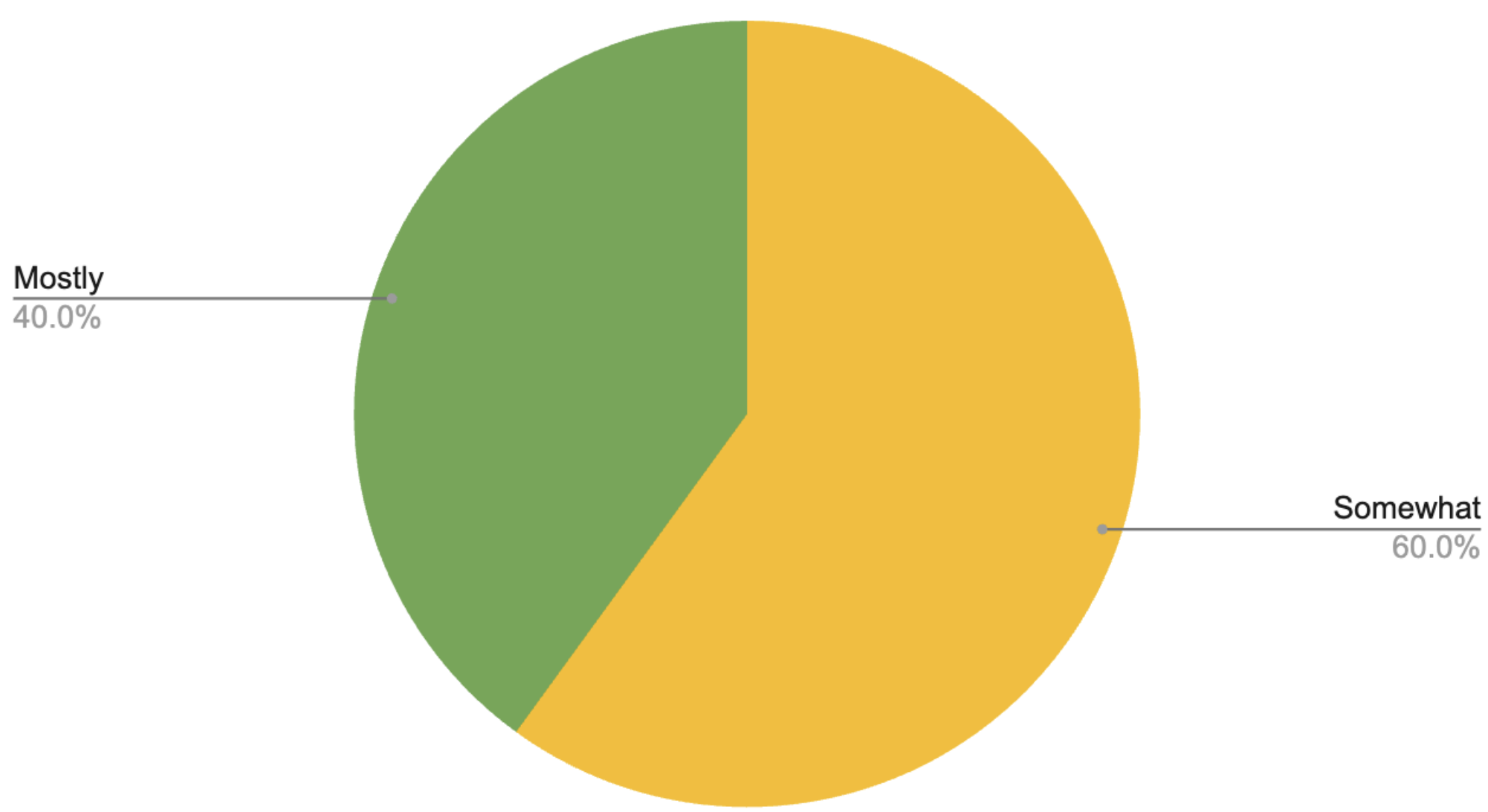

**Figure 4.** Reviewers were asked how much they trusted the pre-review results. The possible responses to this survey question were “Not at all”, “Somewhat”, “Mostly”, and “Completely”. None of the responders chose the first or last response.

Asked which additional criteria they would like the pre-reviewer to cover next year, reviewers most often chose AI use transparency (12) and novelty of approach (11), followed by relevance to BOSC (9), community (7), and suitability for a long talk (4). Opinion here was not unanimous. One reviewer argued against automating the more editorial judgments, such as relevance, community, and suitability, on the grounds that an LLM's recommendation would introduce bias even when offered only as advice, and that such calls are better left to people who can take responsibility for them. When asked whether BOSC should keep offering the AI-assisted pre-review next year, 11 said yes, 5 said maybe, and 1 said no.

The free-text comments in the survey pointed the same way. One reviewer, a co-author of this paper, noted that taking the busywork off reviewers' plates could help us recruit and retain volunteers, and that reviewer valued the help in distinguishing software that is nominally open source from software that can actually be reused in practice. Another described using the pre-review as a cross-check after drafting their own evaluation, going back to it when the two disagreed to see what they might have missed. A third said the tool worked fine but added that they were "in the 'verify anything' stage with AI tools in general". The recurring theme was that reviewers welcomed help with objective, checkable facts and were wary of handing over judgments that need editorial expertise.

# 5. Discussion

The BOSC pre-review process was a good fit for generative AI. In large part, this was thanks to the intellectual labor the BOSC organizers had already done previously of defining the review criteria and creating a scoring rubric for each criterion - something most conferences don't do. Successful use of AI tools typically relies on appropriate crafting of the questions and access to the necessary background material, and BOSC already had set the stage to make that possible.

The pre-reviewer was found to be the most useful for the more objective review criteria (such as "Available"). Confirming the accessibility of a URL, matching a license file against the declared license, tallying commit frequency, and checking a submission against past presentations are all tasks where a simple rule-based check would be likely to miss things, and the AI handled them quickly. Runabilly was especially useful: it automated the build-and-run checks that can take a human reviewer a long time. Running those checks inside a container also kept them from touching the reviewer's own computing environment, helping to reduce risk.

The limitations of the approach resulted, in large part, from the imperfection of the specifications provided to the AI pre-reviewer. Because the AI sticks faithfully to the rubric, it tends to diverge from human judgment where the rubric's authors did not anticipate a case. An example involves licensing. A submission may declare an open license while the repository itself carries no license file, or carries a different one. The AI read this as an unavailable license and assigned a failing score, when the rubric's intended score was medium, a valid license that does not match the declaration. In subtle cases like this, where the strict definition was never written down, the AI sometimes got it wrong, but a human could make the same mistake; the error is better read as imprecision in the rubric than as a failure of the model. The lesson is that getting AI to review well depends less on the model's raw performance or stability than on removing ambiguity from the criteria it is given. Even a criterion that looks mechanically simple can be misled by a submission that falls between the tiers in the rubric, and we cannot prove that the instructions we wrote constitute a perfect rubric. This goes beyond conference review: for any AI-driven judgment, what matters is whether the precision and coverage of the criteria are adequate for the range of inputs the system will see. We also recognize that a single run over 60 abstracts is not enough to generalize from.

Using AI in review raises several ethical questions, which we tried to mitigate by our approach. Critically, we did not let the AI-assisted pre-reviewer assign scores to abstracts or decide whether abstracts should be accepted. Every AI-generated judgment, score, and comment was offered only as optional reference material for human reviewers, and the final scores and decisions were made by people. This approach kept human experts in control of the decisions, and aligned in spirit with ISCB's policy that using generative AI to polish a submission is acceptable while generating the submission text with AI is not. Even so, our finding that AI can do a good deal of the reviewing when the rubric is good enough raises a real question: what is it that humans should be judging in a review? The second issue concerns the privacy of submission data. The records in the EasyChair abstract system can include conflict-of-interest disclosures, requests for financial support, and demographic information. We protected submitters by withholding this sensitive information from the AI, which runs on a third party's servers, and by extracting only the fields the judgment needs before any of it is sent. Ideally we would go further and run a fully open-weight model locally and offline. For a conference built around open source and open data, evaluating open-weight models for this purpose is a natural next step.

As with any application of AI, there are always going to be some doubters or naysayers; two of the BOSC reviewers who took the time to answer the post-review questionnaire reported that they did not use the pre-review results, and one of those felt that we should not run the AI pre-reviewer next year. It is possible that some of the reviewers who did not fill out the survey were of the mindset that AI is bad, wastes resources, and diminishes human satisfaction and trust. The approach we choose for integrating further AI-assisted steps in our process must be carefully crafted to balance human expertise with AI efficiency, and not to alienate those who are not fans of AI tools.

# 6. Conclusion

Conference review, like journal peer review, rests on volunteer labor, but as the number of papers and conference submissions continues to rise rapidly, this paradigm is struggling to keep pace. Generative AI is already changing how research is done, and in information science, including bioinformatics, it is easy to imagine sharp shifts in both the quality and the quantity of output. We felt that shift at BOSC 2026. In that climate, loading still more of the cost onto volunteer reviewers is neither healthy for the community nor sustainable. An approach in which AI supplies verification results and supporting material so that humans can concentrate on judgment will be welcomed by most members of a community that has spent years watching over the development and maintenance of open-source software. The reviewers who used our pre-review process this year appreciated its help assessing objective and/or tedious aspects like the open-source license and the project “runnability”, while wary of surrendering human judgment on more subjective criteria such as “relevance” (which we did not assess with the pre-reviewer). As generative AI continues to improve rapidly, it will be interesting to see what the landscape of submissions looks like in 2027, and how our reviewers’ attitudes about the use of AI to assist in the review process evolves.

# Data and code availability

Both https://github.com/OBF/bosc-pre-review and https://github.com/OBF/runabilly/ are openly available under the BSD 3-Clause license.

# AI-use statement

Generative AI was central to the work this paper reports: our pre-review tools use Claude (through Claude Code, Anthropic's command-line agent) as their reasoning engine, though the approach is model-agnostic and any sufficiently capable agent could serve in its place. We also used Claude to help with analysis and drafting and translating text while preparing this manuscript. However, the authors reviewed and corrected all text as needed, and stand behind the content.

# Acknowledgements

We thank all of the BOSC reviewers for their thoughtful, constructive abstract reviews and for their feedback on the experimental AI-assisted pre-reviewer. The reviewers are listed on https://www.open-bio.org/events/bosc-2026/.

NLH and SC were supported in part by the Director, Office of Science, Office of Basic Energy Sciences, of the U.S. Department of Energy under Contract No. DE-AC02-05CH11231.

# Appendix A: Post-review questionnaire

**BOSC 2026 AI Pre-Review — Reviewer Feedback**

This year, for the first time, we experimented with AI-assisted pre-review. Our goal was to reduce the time reviewers spent checking factual things (e.g. what license an open-source repo uses) and to give reviewers information that could help them in the review process. We would like to know how well that worked.

(* indicates a required question)

Q1. Did you look at the pre-review results? *
[Multiple choice (choose one)]
- No
- A bit
- A lot

----------------------------------------
SECTION: Using the pre-review results
----------------------------------------

Q2. Did you find the pre-review results helpful?
[Multiple choice (choose one)]
- Not at all
- Somewhat
- Very

Q3. How much time do you estimate the pre-review saved you, compared with the time you would have spent without it?
[Multiple choice (choose one)]
- A minute or two per review
- 3–5 minutes per review
- 6–15 minutes per review
- More than 15 minutes per review

Q4. Which pre-review results did you find helpful? (select all that apply)
[Checkboxes (select all that apply)]

Note: These are the checks shown in the pre-review spreadsheet.

- Available — source reachable and has real content
- Open / License — recognized open-source license matching the declared one
- Fresh — whether the project was presented at BOSC in recent years
- Active — whether the project is actively maintained
- Human-generated — whether the submission and the code it's based on appear to be written by a human (rather than AI)
- Runnable — whether the code builds and runs (the “Runabilly” results)

Q5. The “runnable” criterion (whether the software actually builds and runs) was optional to assess. How did you handle it for the abstracts you reviewed? (select all that apply)

[Checkboxes (select all that apply)]

- I tested it by hand myself
- I relied on the pre-review “Runabilly” results
- I did not assess runnability

Q6. Did you notice any cases where the pre-review results were wrong or misleading?

[Multiple choice (choose one)]

- No
- Yes, minor issues
- Yes, significant issues

Q7. If you noticed wrong or misleading results, please describe them.

[Long answer (paragraph)]

Q8. Overall, how much did you trust the pre-review results?

[Multiple choice (choose one)]

- Not at all
- Somewhat
- Mostly
- Completely

Q9. When you used a pre-review result, did you independently double-check it?

[Multiple choice (choose one)]

- Always
- Sometimes
- Rarely
- Never

----------------------------------------

SECTION: Looking ahead

----------------------------------------

Q10. Next year we will probably have the pre-reviewer assess more of the BOSC review criteria (while still leaving the judgements to human reviewers). Which of the remaining criteria would you most appreciate having pre-review results for?

[Checkboxes (select all that apply)]

- Relevant to BOSC
- AI use transparency
- Community
- Novelty of approach
- Suitability for a long talk

Q11. Would you like BOSC to continue offering AI pre-review next year?

[Multiple choice (choose one)]

- No
- Maybe
- Yes

Q12. We'd love to hear any comments you have about the BOSC pre-reviewer.

[Long answer (paragraph)]